\documentclass[dvipsnames]{article} 
\usepackage{ijcai25}

\usepackage{times}
\usepackage{soul}
\usepackage{url}
\usepackage[utf8]{inputenc}
\usepackage[small]{caption}
\usepackage{graphicx}

\usepackage{amsmath}
\usepackage{amsthm}
\usepackage{amsfonts}
\usepackage{algorithm}
\usepackage{algpseudocode}
\DeclareMathOperator*{\argmax}{arg\,max}
\DeclareMathOperator*{\argmin}{arg\,min}

\usepackage{booktabs}
\usepackage[table]{xcolor} 
\usepackage{array}
\usepackage{enumitem}
\usepackage[switch]{lineno}

\definecolor{tabblue}{HTML}{1C77B4}
\definecolor{taborange}{HTML}{FF7F0B}
\definecolor{tabgreen}{HTML}{2AA02B}
\definecolor{tabred}{HTML}{D62729}
\definecolor{tabpurple}{HTML}{9567BC}
\definecolor{customLightGray}{gray}{0.9}

\colorlet{mytextgreen}{black} 
\colorlet{fillgreen1}{OliveGreen!20!white}     
\colorlet{fillgreen2}{OliveGreen!40!white}     
\colorlet{fillgreen3}{OliveGreen!60!white}     
\colorlet{fillgreen4}{OliveGreen!80!white}     
\colorlet{fillgreen5}{OliveGreen}               

\newcommand{\systemname}[1]{Interactive-Reflective Dialogue Alignment}

\usepackage{pgfplots}
\pgfplotsset{compat=1.18}
\usepackage{fontawesome}

\pgfplotsset{
    myStackedBarStyle/.style={
        xbar stacked,
        xmin=0, xmax=100,
        xtick=\empty,
        axis x line=none,
        axis y line=left,
        y axis line style={draw=none},
        ytick style={draw=none},
        bar width=0.7cm,
        width=8cm, 
        nodes near coords={\pgfmathprintnumber[fixed zerofill, precision=1]{\pgfplotspointmeta}\%},
        nodes near coords style={font=\tiny, color=black, anchor=center},
        point meta=explicit,
        ymin=-0.6, ymax=0.6,
        enlarge y limits=false,
        height=2.5cm,
        font=\scriptsize,
        yticklabel style={
            font=\small,
            align=right,
            xshift=-2mm
        }
    }
}

\usepackage[hidelinks]{hyperref}

\title{Reflective Verbal Reward Design for Pluralistic Alignment\footnote{
The behavioral studies in this paper were approved by the University of Waterloo's Human Research Ethics Board (File 46074).
}}

\author{
Carter Blair\(^{\dagger}\)
\and
Kate Larson\and
Edith Law
\affiliations
University of Waterloo\\
\emails
\(^{\dagger}\)cblair@uwaterloo.ca
}

\begin{document}

\maketitle

\begin{abstract}

AI agents are commonly aligned with ``human values'' through reinforcement learning from human feedback (RLHF), where a single reward model is learned from aggregated human feedback and used to align an agent's behavior. However, human values are not homogeneous--different people hold distinct and sometimes conflicting values. Aggregating feedback into a single reward model risks disproportionately suppressing minority preferences. To address this, we present a novel reward modeling approach for learning \textit{individualized} reward models. Our approach uses a language model to guide users through \textit{reflective} dialogues where they critique agent behavior and construct their preferences. This personalized dialogue history, containing the user's reflections and critiqued examples, is then used as context for another language model that serves as an individualized reward function (what we call a ``verbal reward model'') for evaluating new trajectories.
In studies with 30 participants, our method achieved a 9-12\% improvement in accuracy over non-reflective verbal reward models while being more sample efficient than traditional supervised learning methods. 

\end{abstract}

\section{Introduction}

Effectively aligning AI systems with human values requires addressing the inherent diversity and context-dependence of those values \cite{schwartz1992universals,friedman2013value,le2009values}.
However, prevalent alignment techniques, such as RLHF \cite{bai_training_2022}, typically train a \emph{single} reward model from preferences aggregated across multiple users.
By default, these singular models implicitly make trade-offs among competing values and can disproportionately suppress minority viewpoints \cite{siththaranjan2023distributional,chakraborty2024maxmin}.

In response to this shortcoming, recent work has explored approaches that better preserve the diversity of human values \cite{siththaranjan2023distributional,chakraborty2024maxmin,poddar2024personalizing}.
While the specifics differ, these approaches broadly increase the granularity of the reward signal to account for differences in opinion.
For example, some methods learn multiple distinct reward models to capture different preference clusters \cite{chakraborty2024maxmin}.
Others learn reward models that output distributions rather than scalar values to represent uncertainty and variation in preferences \cite{siththaranjan2023distributional}, and others learn \textit{individualized} reward models~\cite{poddar2024personalizing}.

These higher-granularity approaches offer benefits for collective and individual alignment.
In collective settings where we must aggregate preferences across many stakeholders, having granular representations of preferences enables making transparent choices about balancing competing values by explicitly aggregating rewards.
Rather than implicitly averaging preferences, we can optimize for desired properties like egalitarian welfare.
Meanwhile, for personal AI assistants, techniques that learn high granularity preference representations can give rewards that are more representative of the given user compared to the population average.
However, a challenge exists: While a single reward model can leverage data from all users to learn a scalar reward, higher-granularity approaches must learn more complex patterns (e.g., preference distributions \cite{siththaranjan2023distributional} or personalized reward functions \cite{poddar2024personalizing}), which are more challenging to estimate due to their increased dimensionality and sparser per-pattern data coverage.

A further complication arises from the process of \emph{preference construction} in complex and novel domains, where humans actively construct their preferences rather than simply reveal them \cite{lichtenstein2006construction}.
While current reward modeling approaches typically rely on passive data collection and straightforward labeling, this may be insufficient: the process of \emph{constructing} preferences benefits from deliberate reflection, which helps individuals transform their latent values into concrete preferences \cite{fischhoff1991value}.
This insight is supported by findings from psychology, consumer research, and deliberative polling, which demonstrate that prompting people to actively contemplate their underlying values and reasoning processes leads to more well-defined and considered preferences \cite{hauser2014self,ver2020improving,fishkin2005experimenting}.

To address these two challenges, we introduce Interactive-Reflective Dialogue Alignment (IRDA), a system that uses large language models (LLMs) to learn personalized reward functions through interactive dialogue.
IRDA combines three components: (1) \textbf{reflective verbal preference elicitation} that guides users in articulating their values, (2) \textbf{active learning} to strategically select examples for human critique, and (3) \textbf{LLM-driven verbal reward modeling} where the LLM directly serves as the reward function by leveraging its in-context learning ability to generalize from sparse user feedback.
IRDA's architecture confronts the preference construction problem by replacing passive labeling with LLM-guided dialogues that provoke deliberate, context-sensitive reflection (System 2 cognition).
At the same time, its data-efficient learning strategy (active few-shot learning) mitigates the difficulty of learning personalized reward functions.

We evaluated IRDA with two user studies involving 30 participants in total.
The first study (21 participants) focused on building a reward model for each user’s personal definition of ``respectful behavior,'' while the second (9 participants) explored ethical decision-making in autonomous vehicles.
Across both studies, participants differed widely in their value judgments, and our system was able to capture these individual definitions of value-aligned behavior more accurately than baseline approaches.

Our contributions are as follows:
\begin{itemize}
    \item A novel pipeline for aligning AI agents to individual values, informed by AI, HCI, and social science.
    \item A comprehensive evaluation in two distinct domains, demonstrating that our system captures individual preferences more accurately than baselines.
    \item Empirical characterization of how individuals diverge in their conceptions of value-aligned AI behavior.
    \item Insights for future research on interactive systems that help end users construct, refine, and operationalize their latent values.
\end{itemize}

\section{Related Work}

\paragraph{Pluralistic Alignment.} Recent work has highlighted the importance of moving beyond monolithic reward models and toward approaches that capture heterogeneous or uncertain human preferences in AI alignment. For instance, distributional preference learning (DPL) estimates an entire distribution over possible reward values, thereby accommodating hidden context and diverse annotator criteria~\cite{siththaranjan2023distributional}. Similarly, methods like MaxMin-RLHF learn a mixture of reward models and optimize an egalitarian objective to avoid disproportionately favoring majority viewpoints~\cite{chakraborty2024maxmin}. Others have proposed user-specific latent variables that personalize reward models without requiring extensive per-user labels~\cite{poddar2024personalizing} or have leveraged meta-learning to reduce feedback requirements~\cite{iii2022fewshot}. However, these approaches assume users have direct access to their preferences in novel contexts despite evidence from preference elicitation and psychology literature suggesting otherwise. Our work complements these advances by actively eliciting fine-grained, user-specific preferences and helping users turn their latent values into concrete preferences through guided reflection.

\paragraph{Verbal Reward Design.} A separate line of research explores using large language models (LLMs) to act as or generate reward functions. Some methods prompt LLMs to propose reward code, which is then used to train RL policies via standard optimization~\cite{ma2024eureka,xie2024textreward,verma2024balancing,behari2024decision}. Other methods directly treat an LLM as a proxy reward function by prompting it with desired behavior descriptions~\cite{kwon2022reward}. These LLM-based approaches have made reward specification more accessible, particularly in domains where handcrafting objectives is difficult. However, they treat preference specification as a one-way street, where users tell the LLM what they want. Our system makes it a two-way dialogue, using LLMs both to help users clarify their preferences and to translate those preferences into reward functions.

\paragraph{Reflection as a Path to Expressing Latent Preferences.} Preferences are rarely pre-defined artifacts waiting to be extracted; instead, in new contexts, they form through reflective processes that turn latent values into concrete preferences  ~\cite{fischhoff1991value}. In consumer and behavioral research, explicitly prompting users to reflect on trade-offs or alternative perspectives fosters more stable and revealing preference statements~\cite{hauser2014self,ver2020improving}. We build on this insight by weaving reflection into an LLM-based alignment pipeline, enabling users to clarify and externalize their values for AI systems.

\paragraph{Designing Reflection into Dialogue Systems.} Within HCI, frameworks like Fleck and Fitzpatrick’s~\cite{fleck2010reflecting} outline how technologies can scaffold reflective thinking, while studies show that structured prompts—scripted or adaptive—support deeper introspection over time~\cite{kocielnik_reflection_2018,wolfbauer_script_2022,bentvelzen_revisiting_2022}. Recent advances in LLM-driven agents further enable flexible ``reflective dialogues,'' guiding users to articulate emergent values and transform them into actionable preferences~\cite{arakawa2024coaching}. Our contribution is to extend these dialogues to preference elicitation for AI alignment, using LLMs to help users iteratively refine how an AI agent should act.

\section{Interactive-Reflective Dialogue Alignment (IRDA) System}\label{section:irda_system}

We present the \textit{Interactive-Reflective Dialogue Alignment} (IRDA) system, which enables non-expert users to iteratively define a value concept and construct a corresponding reward model for agent training. Our approach is founded on the insight that human values are refined through a process of reflection and iterative feedback \cite{fischhoff1991value}. To this end, IRDA employs a dual-loop framework that first elicits user feedback through a \textcolor{OliveGreen}{\emph{preference construction loop}} over a diversity-based pool of trajectories and then refines the model via an \textcolor{OliveGreen}{\emph{uncertainty reduction loop}} over a separate pool.

To begin, let \(\mathcal{T}_D\) denote the diversity-based pool of trajectories, where each trajectory \(\tau = (s_0,a_0,\dots,s_T)\) represents a sequence of states and actions executed by the agent. We extract features \(\phi(\tau) \in \mathbb{R}^{d}\) from each trajectory and partition \(\mathcal{T}_D\) into \(k\) clusters \(\{\mathcal{C}_1, \ldots, \mathcal{C}_k\}\) using \(k\)-means clustering:
\[
  \{\mathcal{C}_i\}_{i=1}^k = \argmin_{\{\mathcal{C}_i\}} \sum_{i=1}^k \sum_{\tau \in \mathcal{C}_i} \|\phi(\tau) - \mu_i\|_2^2,
\]
with cluster centroids
\[
  \mu_i = \frac{1}{|\mathcal{C}_i|} \sum_{\tau \in \mathcal{C}_i} \phi(\tau).
\]
For each cluster, we select a representative trajectory:
\[
  \tau_i^{cent} = \argmin_{\tau \in \mathcal{C}_i} \|\phi(\tau) - \mu_i\|_2.
\]

In the initial \textcolor{OliveGreen}{\emph{preference construction loop}}, the user specifies a value concept (e.g., “respectfulness”) and provides qualitative feedback \(e_i\) on each centroid trajectory (e.g., ``This is not respectful because...''). Each trajectory is encoded into an ASCII representation \(\alpha(\tau)\), and the resulting feedback is aggregated as
\[
  \mathcal{D}_{fb} = \{(\alpha(\tau_i^{cent}), e_i)\}_{i=1}^k.
\]
Once the feedback is collected, an LLM is queried on the entire \(\mathcal{D}_{fb}\) to generate a hypothesis about what features the user is using to make their decisions, \(\mathcal{H}\), and alternative features the user could consider \(\mathcal{A}\):
\[
  (\mathcal{H}, \mathcal{A}) = G\bigl(\mathcal{D}_{fb}\bigr).
\]
The user is then asked to respond to the generated hypotheses and alternatives, explaining why these features are or are not significant to their decision-making. This is intended to help the user reflect on their values, which can update their mental model \(\mathcal{M}_u\). If \(\mathcal{M}_u\) does change, the user returns to the beginning of the \emph{preference construction loop}.

Once the user confirms that \(\mathcal{M}_u\) is stable, the system transitions to the \textcolor{OliveGreen}{\emph{uncertainty reduction loop}}. In this stage, we consider a separate uncertainty-based pool of trajectories, \(\mathcal{T}_U\). We iteratively refine our reward model on this pool. The reward model is based on an LLM that is prompted to assess whether a trajectory is aligned or not. The LLM is given the entire conversation history \(\mathcal{C}\) (which includes all user feedback, system prompts, and responses), the encoded trajectory \(\alpha(\tau)\), and environment details such as symbol meanings and action spaces (EnvDesc) and outputs token probabilities for labels such as ``respectful'' and ``disrespectful.'' Specifically, the alignment probability $p_\theta(1|\tau)$ and misalignment probability $p_\theta(1|\tau)$ are computed as
\[
  p_\theta(1|\tau), p_\theta(0|\tau)  = f_{\text{LLM}}\bigl(\mathrm{EnvDesc}, \mathcal{C}, \alpha(\tau)\bigr),
\]

where \(p_\theta(1|\tau)\) is the token probability for the ``aligned'' token (e.g., ``respectful'') and \(p_\theta(0|\tau)\) is the token probability for the ``misaligned'' token (e.g., ``disrespectful'').

The associated uncertainty is defined by
\[
  U(\tau) = 1 - \bigl|p_\theta(1|\tau) - p_\theta(0|\tau)\bigr|,
\]
and the trajectory with maximum uncertainty is selected,
\[
  \tau^* = \argmax_{\tau \in \mathcal{T}_U} U(\tau),
\]
and the user is queried for an explanation \(e^*\) regarding \(\tau^*\). This new feedback \((\alpha(\tau^*), e^*)\) is appended to \(\mathcal{D}_{fb}\), and the uncertainty reduction loop is repeated until \(U(\tau^*) < \epsilon\) for a set threshold \(\epsilon\).

The final reward model is an LLM that, using the complete conversation history \(\mathcal{C}\) and all accumulated feedback \(\mathcal{D}_{fb}\), classifies new trajectories as aligned or misaligned with the user’s value. It outputs a binary decision based on token probabilities:
\[
  R(\tau) = \mathbb{I}\bigl[p_\theta(1|\tau) > p_\theta(0|\tau)\bigr].
\]

\begin{algorithm}[t]
\caption{\systemname{}}
\label{alg:dual_loop}
\begin{algorithmic}[1]
\State \textbf{Input:} $\mathcal{T}_D$, $\mathcal{T}_U$, value $v$, threshold $\epsilon$, EnvDesc
\State \textbf{Preprocessing:} For each $\tau\in\mathcal{T}_D$, extract $\phi(\tau)$; cluster via $k$-means; select representatives $\{\tau_i^{cent}\}$.
\Repeat \hfill \textcolor{OliveGreen}{\textit{Preference Construction Loop}}
    \State For each $\tau_i^{cent}$, obtain/update label $e_i$; form 
    \[
      \mathcal{D}_{fb} = \{(\alpha(\tau_i^{cent}), e_i)\}.
    \]
    \State Query the LLM with $\mathcal{D}_{fb}$ to yield feature hypotheses $\mathcal{H}$ and alternatives $\mathcal{A}$.
    \State User responds to $(\mathcal{H},\mathcal{A})$ and refines $\mathcal{M}_u$. 
\Until{User confirms $\mathcal{M}_u$ is stable}
\Repeat \hfill \textcolor{OliveGreen}{\textit{Uncertainty Reduction Loop}}
    \For{each $\tau\in\mathcal{T}_U$}
        \State $p_\theta(1|\tau), p_\theta(0|\tau) = f_{\text{LLM}}(\mathrm{EnvDesc},\mathcal{C},\alpha(\tau))$
        \State $U(\tau) = 1 - \left|p_\theta(1|\tau)-p_\theta(0|\tau)\right|$
    \EndFor
    \State $\tau^* = \argmax_{\tau\in\mathcal{T}_U} U(\tau)$.
    \State Query the user for $e^*$ on $\tau^*$; update 
    \[
      \mathcal{D}_{fb} \gets \mathcal{D}_{fb} \cup \{(\alpha(\tau^*), e^*)\}.
    \]
\Until{$U(\tau^*) < \epsilon$}
\State \textbf{Final Reward Model:} For any new $\tau$, define 
\[
  R(\tau) = \mathbb{I}\bigl[p_\theta(1|\tau) > p_\theta(0|\tau)\bigr],
\]
\State \textbf{Return:} $R(\cdot)$.
\end{algorithmic}
\end{algorithm}

This process, beginning with diversity-based sampling, followed by preference construction, and culminating in uncertainty-driven refinement, yields a final LLM-based reward model. This final LLM-based reward model uses the entire conversation history and all explained examples. The full process is formalized in Algorithm \ref{alg:dual_loop}.

\section{Study Design \& Methodology}
We evaluated our system (implemented with \texttt{gpt-3.5}) in two studies: Study 1 investigates the utility of our system for learning about participants' definition of \textit{respectful} agent behavior. Study 2 investigates the utility of our system for learning about participants' decision-making in moral dilemmas involving an agent (autonomous vehicle). Our studies employ a within-subject design, collecting data from each participant to train and test each method. Our studies aim to answer the following three questions:

\begin{quote}
\begin{enumerate}[label=\textbf{RQ\arabic*:}]
    \item How do individuals' interpretations of value-aligned AI behavior differ?
    \item Does structured reflection enhance verbal reward modeling?
    \item When is individualized verbal reward modeling effective?
\end{enumerate}
\end{quote}

\subsection{Environments}

\paragraph{Multi-Agent Apple Farming Environment (Study 1).} A $6\times6$ grid contains apples and garbage, with one ``main'' (blue) agent and three ``background'' (gray) agents. Each agent ``owns'' one of four $3\times3$ orchards. Two background agents remain stationary, and one moves freely. The main agent is rewarded by picking apples (none for collecting garbage). Participants assess whether the blue agent behaves ``respectfully.''

\paragraph{Moral Machine Environment (Study 2).} Adapted from \cite{awad2018moral}, this environment features ethical dilemmas in which an autonomous vehicle must stay on course or swerve, potentially harming different combinations of pedestrians or passengers (including children, adults, the elderly, and animals). Each outcome varies in factors like legality, social status, and species. Participants decide which outcome the car should choose.

\subsection{Participants}
\begin{figure}[tbhp]
    \centering 
    \begin{tikzpicture}[every node/.style={font=\small}]

    \begin{axis}[
        myStackedBarStyle, 
        name=study1axis, 
        ytick={0},       
        yticklabels={Study 1},
        font=\small,
    ]

    \addplot [fill=fillgreen1, draw=none] coordinates {(9.5,0) [9.5]};
    \addplot [fill=fillgreen2, draw=none] coordinates {(42.9,0) [42.9]}; 
    \addplot [fill=fillgreen3, draw=none] coordinates {(38.1,0) [38.1]}; 
    \addplot [fill=fillgreen4, draw=none] coordinates {(9.5,0) [9.5]}; 
    \end{axis}

    \begin{axis}[
        myStackedBarStyle,
        name=study2axis,
        at={(study1axis.south)}, 
        anchor=north,        
        yshift=-3mm,         
        ytick={0},           
        yticklabels={Study 2},
        font=\footnotesize
    ]

    \addplot [fill=fillgreen2, draw=none] coordinates {(22.2,0) [22.2]};
    \addplot [fill=fillgreen3, draw=none] coordinates {(33.3,0) [33.3]};
    \addplot [fill=fillgreen4, draw=none] coordinates {(11.1,0) [11.1]};
    \addplot [fill=fillgreen5, draw=none] coordinates {(33.3,0) [33.3]};
    \end{axis}

    \end{tikzpicture}
    \caption{Participant familiarity with reinforcement learning for Study 1 (top) and Study 2 (bottom). Key: \textcolor{fillgreen1}{\protect\rule{0.75em}{0.75em}}~Very Unfamiliar; \textcolor{fillgreen2}{\protect\rule{0.75em}{0.75em}}~Unfamiliar; \textcolor{fillgreen3}{\protect\rule{0.75em}{0.75em}}~Neutral; \textcolor{fillgreen4}{\protect\rule{0.75em}{0.75em}}~Familiar; \textcolor{fillgreen5}{\protect\rule{0.75em}{0.75em}}~Very Familiar.}
    \label{fig:combined_familiarity} 
\end{figure}
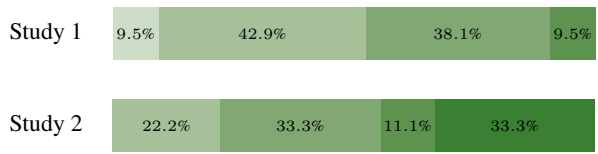

In Study 1, we recruited 21 participants from the University of Waterloo (18 to 39 age range, M=23.86, 7 self-identified as male and 14 as female). When asked to rate their level of familiarity with reinforcement learning on a 5-point Likert ranging from ``very unfamiliar'' (1) to ``very familiar'' (5), the mean level of familiarity was 2.48, with the mode and median being 2. The Likert-scale data for Study 1 (Figure~\ref{fig:combined_familiarity}, top bar) highlights that more than half of participants were ``unfamiliar'' or ``very unfamiliar'' with reinforcement learning.

In Study 2, we recruited 9 participants from the University of Waterloo (18 to 33 age range, M=25.66, 6 self-identified as male and 3 as female). When asked to rate their level of familiarity with reinforcement learning on a 5-point Likert scale ranging from ``very unfamiliar'' to ``very familiar,'' the mean level of familiarity was 3.55, with the mode and median being 3. The Likert-scale data for Study 2 is visualized in Figure~\ref{fig:combined_familiarity} (bottom).

\subsection{Procedure} 
Participants used our system to specify how they would like the agent to act. Participants then labeled 50 scenarios and were interviewed.

\textit{Introduction ($\sim$5min)} - After participants completed the consent form and demographic questions, we thoroughly explained the environment mechanics so that differences reflected their opinions rather than assumptions about the setup.

\textit{Dialogue} - The participant began by conversing with the system about the agent's behavior following the process described in Section~\ref{section:irda_system}. To control the amount of time users spent, we limited the user to one preference construction loop and one uncertainty reduction loop. 

\textit{Labeling} - Following the participants' dialogue interaction with the system, participants labeled 50 scenarios. Each participant labeled the same scenarios, which allowed us to assess how much the participants agreed on the labels.

\textit{Semi-structured Interview ($\sim$10min)} - After completing the labeling task, participants were asked about their ability to communicate their decision-making to the system, including their capacity to articulate label choices, any difficulties in decision-making, and potential changes in their labeling behavior over time.

\subsection{Baseline Comparisons}

We compared IRDA to another verbal reward design system and various supervised learning methods.

\subsubsection{Verbal Baseline ($L^B$)}\label{baseline}
Kwon et al. \shortcite{kwon2022reward} proposed a reward modeling pipeline for text-based environments where the user selects multiple examples from a palette of examples of the agent behaving as they would desire, accompanied by explanations. We modify their pipeline in the following way: Instead of asking the user to select examples from a handcrafted palette, we choose the examples the user sees with the diversity- and uncertainty-based sampling procedures described and in~\autoref{section:irda_system}. This system differs from IRDA in that it does not engage the user in reflective dialogue.

\subsubsection{Supervised Learning Baselines}
We evaluated our approach against neural network-based supervised learning methods, including both individual models per participant and collective models trained on aggregated data. In Study 1, we implemented multi-layer perceptron models (one hidden layer, 32 neurons) using a tensor encoding of the trajectory: individual models ($\text{MLP}^{\text{ind}}_i$) for each participant $i$ and a collective model ($\text{MLP}^{\text{col}}$) trained on all participant data. Study 2 expanded this comparison to include both MLP models with 26-dimensional Moral Machine scenario vectors and convolutional neural networks ($\text{CNN}^{\text{ind}}_i$ and $\text{CNN}^{\text{col}}$) using scenario image inputs. The CNNs used two convolutional layers (16 and 32 filters) with max pooling, followed by fully connected layers reducing to 64 dimensions. Each supervised learning model was incrementally trained with up to 30 examples \textit{per participant} using the Adam optimizer (learning rate 0.001).

\subsection{Analysis}

To answer our research questions, we employ a mixed-methods approach, combining quantitative analyses of model performance and inter-annotator agreement with qualitative analyses of participant decision-making processes and experiences. The mapping between research questions and analysis methods is shown in Table 1.

\begin{table}[t]
  \centering
  
  \label{tab:analysis-rq-mapping}
  \rowcolors{2}{customLightGray}{white}
  \begin{tabular}{%
    >{\raggedright\arraybackslash}m{0.66\columnwidth} 
    >{\centering\arraybackslash}m{0.05\columnwidth}    
    >{\centering\arraybackslash}m{0.05\columnwidth}    
    >{\centering\arraybackslash}m{0.055\columnwidth}    
  }
    \toprule
    \textbf{Analysis Method} & RQ1 & RQ2 & RQ3 \\
    \midrule
    Inter-Annotator Agreement                 & $\checkmark$ &             & $\checkmark$ \\
    Evaluation of Verbal Reward Model Performance  &             & $\checkmark$ &             \\
    Comparison to Supervised Learning                & $\checkmark$ &             & $\checkmark$ \\
    Qualitative Analysis of Participant Decision Making   & $\checkmark$ &             &             \\
    Analysis of Feature Similarity Between Participants  & $\checkmark$ &             & $\checkmark$ \\
    Interview Thematic Analysis              &             & $\checkmark$ & $\checkmark$ \\
    \bottomrule
  \end{tabular}
  \caption{Mapping of Analysis Methods to Research Questions}
\end{table}

\subsubsection{Inter-Annotator Agreement}\label{section:inter_ann_agree}
We assess the inter-annotator agreement between participants on the test set of scenarios they labeled in each study. Since each participant labeled the same test scenarios, we can use Fleiss' kappa value to quantify the inter-annotator agreement between the participants \cite{landis1977measurement}. Generally, kappa statistics below 0 indicate ``poor'' agreement and kappa statistics above 0.8 indicate ``nearly perfect'' agreement \cite{landis1977measurement}.

\subsubsection{Evaluation of Verbal Reward Model Performance}\label{section:eval_lbrm}
We evaluated our system against a baseline without dialogic reflection using a performance metric $P$, where $P_i^\text{IRDA}$ and $P_i^B$ represent participant $i$'s metrics for our system and baseline, respectively. Study 1 used balanced accuracy due to class imbalance, while Study 2 used accuracy.
For each participant, both systems generated rewards for 20 non-training scenarios, yielding 20 pairs of $P$ values. We conducted three statistical tests on the $P$ values:
\begin{enumerate}
    \item We bootstrapped 95\% confidence intervals for the mean by resampling 10,000 times with replacement.
    \item For each participant, we calculated the difference $\Delta P_i = P_i^\text{IRDA} - P_i^B$ and bootstrapped these differences in the same way.
    \item $P$ values were compared using the Wilcoxon signed-rank test, chosen for its robustness to non-normal distributions and reduced false positives \cite{bridge1999increasing}.
\end{enumerate}

\subsubsection{Comparison to Supervised Learning}\label{section:comp_sl}

We compared our verbal system (IRDA) to traditional supervised learning approaches. Both the individual models ($\text{MLP}^{\text{ind}}_i$ and $\text{CNN}^{\text{ind}}_i$) and the collective models ($\text{MLP}^{\text{col}}$ and $\text{CNN}^{\text{col}}$) were trained incrementally, gradually increasing the number of samples used per participant. This methodology allowed us to analyze how model performance evolved with increasing data availability.
For each increment, we calculated $P^\text{ind}_i$ and $P^\text{col}_i$ for each participant $i$. To ensure robustness, we bootstrapped these values with replacement using 10,000 resamples.

\subsubsection{Qualitative Analysis of Participant Decision Making}\label{section:qual_decision}
We conducted a detailed analysis of the message exchanges between participants and the system to gain insight into participants' decision-making processes. We employed an inductive coding approach, systematically reviewing the messages to identify key features and criteria that participants used in their decision-making. Our coding process involved multiple passes through the data, with iterative refinement of the codebook to ensure it captured the full range of decision-making strategies observed.

\subsubsection{Analysis of Feature Similarity Between Participants}\label{section:feature_similarity}
To quantify how similar participants were in their use of decision-making features, we employed the Jaccard similarity coefficient. This measure calculates the overlap between two sets of items, which, in our case, are features the two participants used to make decisions~\cite{jaccard1912distribution}. We computed the Jaccard similarity coefficient for every pair of participants, using the set of decision-making features each participant employed (as identified in our qualitative analysis). To estimate the overall similarity across our participant pool, we then calculated the mean of these pairwise Jaccard coefficients. We used bootstrapping with 10,000 resamples to determine the 95\% confidence intervals.

\subsubsection{Thematic Analysis of Interview Data}\label{section:thematic}
We conducted semi-structured interviews with participants to understand their experiences. The interview transcripts were analyzed using a thematic analysis approach guided by the principles outlined by Braun and Clarke \shortcite{braun2006using}. We followed a six-phase process: familiarization with the data, generating initial codes, searching for themes, reviewing themes, defining and naming themes, and producing the report.

\section{Results: Study 1 - Multi-Agent Apple Farming}

On average, participants took 15 minutes 57 seconds (SD $=$ 6 min. 43 sec., range: 6 min. 59 sec. - 30 min. 55 sec.) to complete the dialogue with the system and 13 minutes 37 seconds (SD $=$ 3 min. 2 sec., range: 6 min. 55 sec. - 18 min. 26 sec.) to complete the labeling of 50 trajectories. Of 21 participants, 7 (33.3\%) entered the \textit{preference construction loop} for one iteration.

\subsubsection{S1 -- Inter-Annotator Agreement}
We observed a Fleiss' kappa value between all participants' labels on the 50 labeled trajectories of $\kappa = 0.336$, indicating ``fair'' agreement among participants \cite{landis1977measurement}. The Fleiss' kappa statistic of 0.336 we observed lends credence to the idea that human values and preferences are subjective and personal.

\subsubsection{S1 -- Evaluation of Verbal Reward Model Performance}
On average, the reward models produced by IRDA achieved significantly higher balanced accuracy scores (measured in percentages) than the baseline system ($L^B$) by 9\% (95\% CI: [5\%, 13\%], M = $68\%$ vs. M = $59\%$, p=.002). This indicates that structured reflection is beneficial.

\subsubsection{S1 -- Comparison to Supervised Learning}

\begin{figure}[t]
  \centering
  \includegraphics[width=0.99\linewidth]{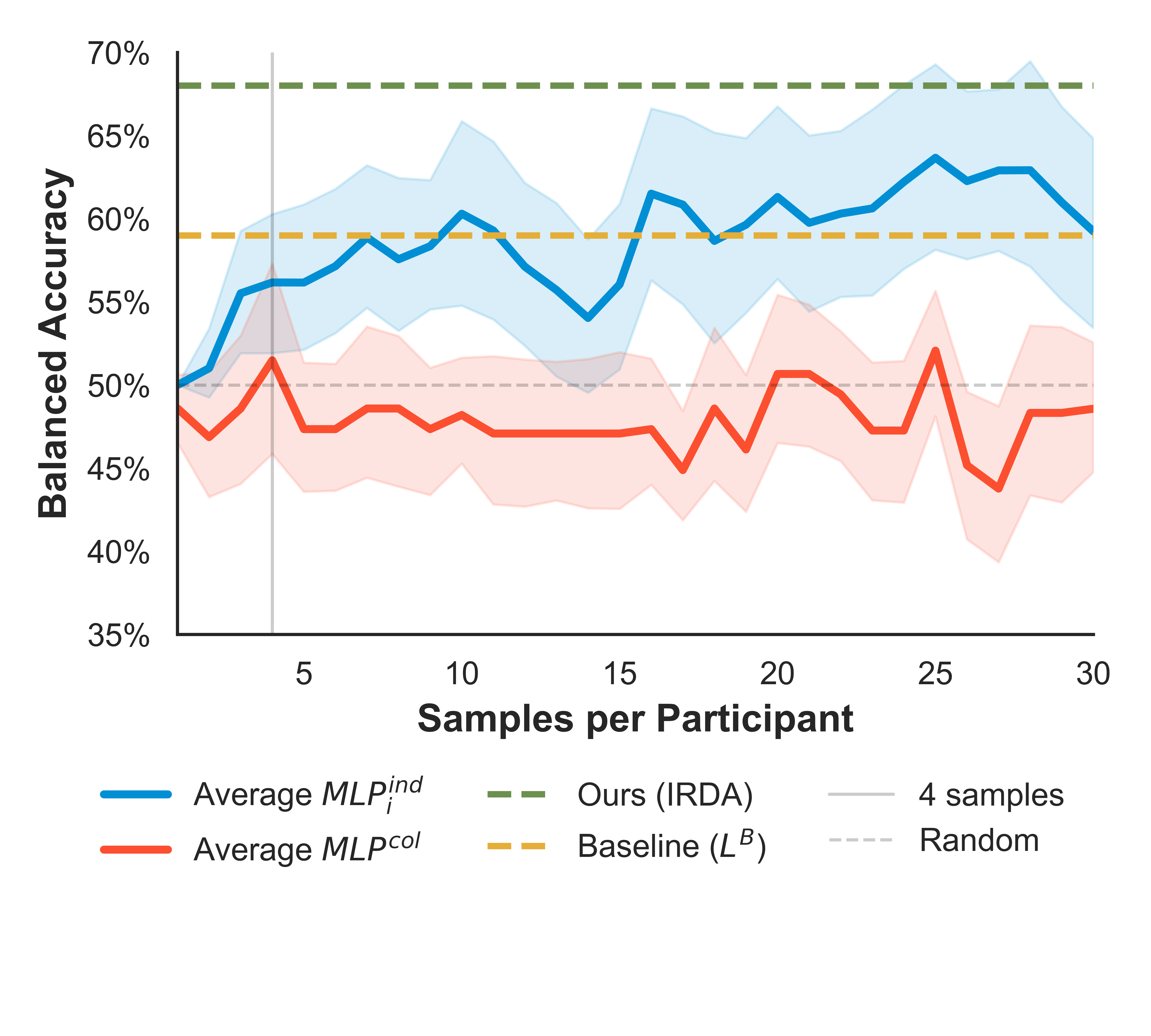}
  \caption{Balanced accuracy of models vs. samples per participant in \textbf{Study 1}. Blue line shows average individual MLP (MLP\textsuperscript{ind}); red shows collective MLP (MLP\textsuperscript{col}). IRDA (green dashed) and baseline ($L^B$, yellow dashed) used four samples per participant (vertical gray line). The collective model was trained on 21 times the samples shown (21 participants). Shaded areas: 95\% confidence intervals.}

  \label{SL_s1}
\end{figure}

With all 30 training samples, the average balanced accuracy of the individual models ($\text{MLP}^\text{ind}_i$) was 59\% (95\% CI: [53\%, 65\%]) while the collective model ($\text{MLP}^{\text{col}}$) achieved 48\% (95\% CI: [46\%, 50\%]). This indicates that participant value definitions varied widely.
\autoref{SL_s1} illustrates the relationship between model performance and the number of samples provided per participant.

\subsubsection{S1 -- Qualitative Analysis of Participant Decision Making}\label{section:s1_features}

Although our system can align AI agents with various values, we focused on respect to examine how individuals interpret even a single value differently. Analysis of participant conversations revealed 12 distinct behavioral features used to evaluate respectful agent behavior (see supplementary material for a list and definitions). Usage varied widely - P1 relied solely on whether agents stayed in their quadrant, while P5, P7, and P10 each employed seven features. Only one participant pair shared identical feature sets, and most participants combined them using hierarchical and conditional rules. While the agent staying in its quadrant was the most common feature, features varied in temporal scope, from static properties (current quadrant location) to multi-step sequences (collecting garbage before apples).

\subsubsection{S1 -- Analysis of Feature Similarity Between Participants}

We observed an average Jaccard similarity coefficient between all pairs of participants' feature usage of $J = 0.357$, with a 95\% confidence interval of $(0.333, 0.3813)$.

\subsubsection{S1 -- Thematic Analysis of Interview Data} Our thematic analysis revealed two main themes: participants' evolving definitions of respect and the system's impact on this evolution.

\textit{\textbf{Evolving Definitions of Respect.}}
Participants' understanding of respect developed through system interaction. Initial definitions focused on simple concepts like spatial boundaries and task-specific behaviors (P4, P6, P7, P10, P19). Through engagement with examples and system feedback, these views became more nuanced. Participants who encountered examples challenging their initial perspectives often expanded their conceptualization of respect (P3, P8, P10, P13, P18, P19, P20). This evolution aligns with consumer research showing that reflection and realistic decision-making improve preference reporting~\cite{hauser2014self}.

\textit{\textbf{Specific Impact of System Hypothesis.}}
The system's hypotheses and alternative features actively shaped participants' respect definitions. The presentation of alternatives prompted reevaluation and revision of initial concepts (P3, P8, P13, P18, P20). While some participants maintained their original views (e.g., P19), the system's suggestions helped others refine their understanding (P13, P18) or consider new perspectives (P3, P20), demonstrating the value of reflective dialogue.

\section{Results: Study 2 - The Moral Machine}

On average, participants took 18 minutes 28 seconds (SD $=$ 7 min. 15 sec., range: 12 min. 0 sec. - 34 min. 34 sec.) to complete the dialogue with the system and 11 minutes 51 seconds (SD $=$ 4 min. 15 sec., range: 4 min. 46 sec. - 17 min. 04 sec.) to complete the labeling of 50 trajectories. Of 9 participants, 1 (11.1\%) entered the \textit{preference construction loop} for one iteration.

\subsubsection{S2 -- Inter-Annotator Agreement}
We observed a Fleiss' kappa value between all participants' labels on the 50 labeled trajectories of $\kappa = 0.460$, indicating ``moderate'' (higher than ``fair'') agreement among participants \cite{landis1977measurement}.

\subsubsection{S2 -- Evaluation of Verbal Reward Model Performance}
On average, the reward models produced by IRDA received higher accuracy scores (measured in percentages) than the baseline system ($L^B$) by 12\% (95\% CI: [4\%, 27\%], M = $65\%$ vs. M = $53\%$, p=.05). This adds more evidence in favor of the effectiveness of structured reflection.

\subsubsection{S2 -- Comparison to Supervised Learning}

\begin{figure}
  \centering
  \includegraphics[width=0.99\linewidth]{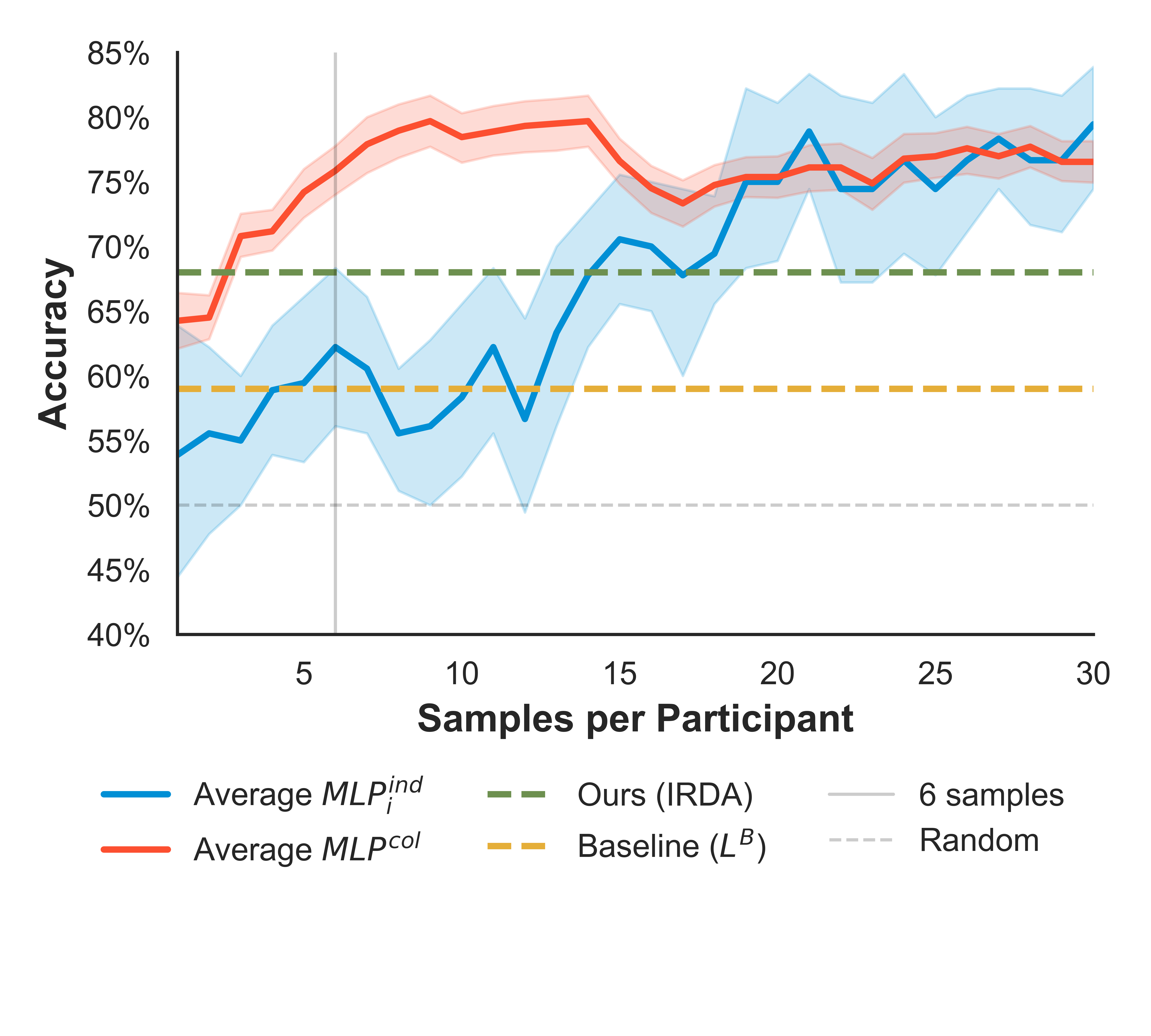}
  \caption{Model accuracies vs. samples per participant in \textbf{Study 2}. Blue: average individual MLP (MLP\textsuperscript{ind}); red: collective MLP (MLP\textsuperscript{col}); green dashed: our IRDA approach; yellow dashed: baseline $L^B$. Shaded areas show confidence intervals. Vertical gray line: 6-sample training point for IRDA and $L^B$. The collective model used nine times the samples shown (9 participants).}
  \label{fig:sl_s2_left}
\end{figure}

With 30 training samples, individual MLPs achieved 79\% accuracy (95\% CI: [74\%, 84\%]) while the collective MLP reached 77\% (95\% CI: [75\%, 78\%]). For CNNs, individual models achieved 67\% accuracy (95\% CI: [61\%, 73\%]) while the collective model ($\text{CNN}^{\text{col}}$) reached 77\% (95\% CI: [70\%, 83\%]). \autoref{fig:sl_s2_left} shows performance versus sample count. These results suggest that collective methods excel with high participant agreement, particularly for complex learning problems like CNN-based image processing, where sample pooling proves beneficial.

\subsubsection{S2 -- Qualitative Analysis of Participant Decision Making}\label{section:s2_features}

By analyzing the participants' conversations with our system, we identified nine features they used in their decision-making in Study 2 (see supplementary material for a list and definitions). The most common features were minimizing casualties (8/9 participants) and traffic rule compliance (8/9 participants). Most participants combined features conditionally and hierarchically.

\subsubsection{S2 -- Analysis of Feature Similarity Between Participants}

We observed an average Jaccard similarity coefficient between all pairs of participants' feature usage of $J = 0.464$, with a 95\% confidence interval of $(0.403, 0.526)$.

\subsubsection{S2 -- Thematic Analysis of Interview Data}Through our thematic analysis, we found two main themes: (1)
Participants’ definitions of ethical decision making evolved throughout the activity, and (2) participants' decisions were primarily based on explicit reasoning but sometimes relied on intuition.

\textit{\textbf{Decision-making Evolution. }}
Participants' decision-making evolved differently during the study. Some expanded their criteria as they encountered more scenarios (P3: ``realized the need to consider new factors when initial factors were equal''), reinforcing Study 1's findings about preference evolution. Others maintained consistent frameworks throughout (P2: ``rules remained consistent'').

\textit{\textbf{Intuition vs. Explicit Reasoning. }}
While most participants could articulate clear reasoning, some relied on intuition for complex scenarios. P7 reported using ``first instinct'' or ``vibes'' for several challenging cases.

\section{Discussion}
Our results suggest that value diversity exists in the context of AI alignment and that verbal reward modeling with LLM-guided reflection can be an effective method for learning individualized reward models that preserve this value diversity.

\paragraph{Value Diversity (RQ1).}
In Study 1, focusing on respectful behavior, we observed substantial diversity in how participants defined and evaluated respect. The low inter-annotator agreement ($\kappa = 0.336$) and feature similarity ($J = 0.357$) suggest fundamental differences in value interpretation rather than noise. This was further evidenced by the stark performance gap between individual and collective models - while individual models achieved 59\% accuracy with just 30 samples, collective models failed to surpass random performance despite access to 21 times more data.

Study 2, examining ethical decisions in autonomous vehicles, revealed more homogeneous preferences ($\kappa = 0.460$, $J = 0.464$). Here, collective models outperformed individual approaches. This contrast between studies highlights how value diversity varies by context and challenges the assumption that universal values can be embedded in AI systems.

\paragraph{The Effectiveness of Reflection (RQ2).}
Our reflection-based approach was effective in both contexts, outperforming baseline methods even when participants held fixed opinions. This suggests that structured reflection enhances preference communication regardless of preference malleability, aligning with dual-process theories in cognitive psychology that suggest reflection can enhance articulation~\cite{evans2019reflections}. Given this, we suggest preference elicitation requires more sophisticated interaction paradigms than simple feedback collection and should engage users in reflection.

\paragraph{Contextual Efficacy of Verbal RMs (RQ3).}
The effectiveness of our approach varies with context: it excels with heterogeneous preferences and limited samples but becomes less critical when preferences are homogeneous and data is abundant. This suggests a complementary approach to value alignment, where methods are selected based on preference heterogeneity and data availability.

\paragraph{Limitations and Future Work.} Our participant pool was relatively homogeneous, so results may not generalize to broader populations or values. Additionally, repeatedly querying an LLM at training time can become prohibitively expensive, limiting scalability for large-scale RL tasks. An alternative is having the LLM produce a standalone code-based reward function, which is more computationally efficient but less flexible. Future work should evaluate this approach in more diverse settings (including LLM alignment tasks) and develop principled methods for aggregating individualized reward models. Social choice offers one promising path for navigating these aggregation decisions~\cite{pmlr-v235-conitzer24a}. In sum, our findings highlight the importance of reflection-based elicitation for capturing preferences and underscore the need for methods designed to accommodate genuine value pluralism in AI alignment.

\section{Open Science and Reproducibility}

All stimuli, a list of identified behavioral features with descriptions, and prompts are available in \href{https://osf.io/8yxf2/?view_only=4e17da7cf0f74bda9da166c33fa151ad}{our project's OSF repository}.

\section*{Acknowledgments}

This work was supported by NSERC Discovery Grants RGPIN-03402-2023 and RGPIN-50503-11794.

\bibliographystyle{named}
\bibliography{ijcai25}

\begin{thebibliography}{}

\bibitem[\protect\citeauthoryear{Arakawa and Yakura}{2024}]{arakawa2024coaching}
Riku Arakawa and Hiromu Yakura.
\newblock Coaching copilot: blended form of an llm-powered chatbot and a human coach to effectively support self-reflection for leadership growth.
\newblock In {\em Proceedings of the 6th ACM Conference on Conversational User Interfaces}, pages 1--14, 2024.

\bibitem[\protect\citeauthoryear{Awad \bgroup \em et al.\egroup }{2018}]{awad2018moral}
Edmond Awad, Sohan Dsouza, Richard Kim, Jonathan Schulz, Joseph Henrich, Azim Shariff, Jean-Fran{\c{c}}ois Bonnefon, and Iyad Rahwan.
\newblock The moral machine experiment.
\newblock {\em Nature}, 563(7729):59--64, 2018.

\bibitem[\protect\citeauthoryear{Bai \bgroup \em et al.\egroup }{2022}]{bai_training_2022}
Yuntao Bai, Andy Jones, Kamal Ndousse, Amanda Askell, Anna Chen, Nova DasSarma, Dawn Drain, Stanislav Fort, Deep Ganguli, Tom Henighan, Nicholas Joseph, Saurav Kadavath, Jackson Kernion, Tom Conerly, Sheer El-Showk, Nelson Elhage, Zac Hatfield-Dodds, Danny Hernandez, Tristan Hume, Scott Johnston, Shauna Kravec, Liane Lovitt, Neel Nanda, Catherine Olsson, Dario Amodei, Tom Brown, Jack Clark, Sam McCandlish, Chris Olah, Ben Mann, and Jared Kaplan.
\newblock Training a {Helpful} and {Harmless} {Assistant} with {Reinforcement} {Learning} from {Human} {Feedback}, April 2022.
\newblock arXiv:2204.05862 [cs].

\bibitem[\protect\citeauthoryear{Behari \bgroup \em et al.\egroup }{2024}]{behari2024decision}
Nikhil Behari, Edwin Zhang, Yunfan Zhao, Aparna Taneja, Dheeraj Nagaraj, and Milind Tambe.
\newblock A decision-language model (dlm) for dynamic restless multi-armed bandit tasks in public health.
\newblock {\em arXiv preprint arXiv:2402.14807}, 2024.

\bibitem[\protect\citeauthoryear{Bentvelzen \bgroup \em et al.\egroup }{2022}]{bentvelzen_revisiting_2022}
Marit Bentvelzen, Paweł~W. Woźniak, Pia~S.F. Herbes, Evropi Stefanidi, and Jasmin Niess.
\newblock Revisiting {Reflection} in {HCI}: {Four} {Design} {Resources} for {Technologies} that {Support} {Reflection}.
\newblock {\em Proceedings of the ACM on Interactive, Mobile, Wearable and Ubiquitous Technologies}, 6(1):2:1--2:27, March 2022.

\bibitem[\protect\citeauthoryear{Braun and Clarke}{2006}]{braun2006using}
Virginia Braun and Victoria Clarke.
\newblock Using thematic analysis in psychology.
\newblock {\em Qualitative Research in Psychology}, 3(2):77--101, 2006.

\bibitem[\protect\citeauthoryear{Bridge and Sawilowsky}{1999}]{bridge1999increasing}
Patrick~D Bridge and Shlomo~S Sawilowsky.
\newblock Increasing physicians’ awareness of the impact of statistics on research outcomes: comparative power of the t-test and wilcoxon rank-sum test in small samples applied research.
\newblock {\em Journal of Clinical Epidemiology}, 52(3):229--235, 1999.

\bibitem[\protect\citeauthoryear{Chakraborty \bgroup \em et al.\egroup }{2024}]{chakraborty2024maxmin}
Souradip Chakraborty, Jiahao Qiu, Hui Yuan, Alec Koppel, Furong Huang, Dinesh Manocha, Amrit Bedi, and Mengdi Wang.
\newblock Maxmin-rlhf: Towards equitable alignment of large language models with diverse human preferences.
\newblock In {\em ICML 2024 Workshop on Models of Human Feedback for AI Alignment}, 2024.

\bibitem[\protect\citeauthoryear{Conitzer \bgroup \em et al.\egroup }{2024}]{pmlr-v235-conitzer24a}
Vincent Conitzer, Rachel Freedman, Jobst Heitzig, Wesley~H. Holliday, Bob~M. Jacobs, Nathan Lambert, Milan Mosse, Eric Pacuit, Stuart Russell, Hailey Schoelkopf, Emanuel Tewolde, and William~S. Zwicker.
\newblock Position: Social choice should guide {AI} alignment in dealing with diverse human feedback.
\newblock In Ruslan Salakhutdinov, Zico Kolter, Katherine Heller, Adrian Weller, Nuria Oliver, Jonathan Scarlett, and Felix Berkenkamp, editors, {\em Proceedings of the 41st International Conference on Machine Learning}, volume 235 of {\em Proceedings of Machine Learning Research}, pages 9346--9360. PMLR, 21--27 Jul 2024.

\bibitem[\protect\citeauthoryear{Evans}{2019}]{evans2019reflections}
Jonathan St~BT Evans.
\newblock Reflections on reflection: The nature and function of type 2 processes in dual-process theories of reasoning.
\newblock {\em Thinking \& Reasoning}, 25(4):383--415, 2019.

\bibitem[\protect\citeauthoryear{Fischhoff}{1991}]{fischhoff1991value}
Baruch Fischhoff.
\newblock Value elicitation: Is there anything in there?
\newblock {\em American Psychologist}, 46(8):835, 1991.

\bibitem[\protect\citeauthoryear{Fishkin and Luskin}{2005}]{fishkin2005experimenting}
James~S Fishkin and Robert~C Luskin.
\newblock Experimenting with a democratic ideal: Deliberative polling and public opinion.
\newblock {\em Acta politica}, 40:284--298, 2005.

\bibitem[\protect\citeauthoryear{Fleck and Fitzpatrick}{2010}]{fleck2010reflecting}
Rowanne Fleck and Geraldine Fitzpatrick.
\newblock Reflecting on reflection: framing a design landscape.
\newblock In {\em Proceedings of the 22nd Conference of the Computer-Human Interaction Special Interest Group of Australia on Computer-Human Interaction}, pages 216--223, 2010.

\bibitem[\protect\citeauthoryear{Friedman \bgroup \em et al.\egroup }{2013}]{friedman2013value}
Batya Friedman, Peter~H Kahn, Alan Borning, and Alina Huldtgren.
\newblock Value sensitive design and information systems.
\newblock {\em Early Engagement and New Technologies: Opening Up the Laboratory}, pages 55--95, 2013.

\bibitem[\protect\citeauthoryear{Hauser \bgroup \em et al.\egroup }{2014}]{hauser2014self}
John~R Hauser, Songting Dong, and Min Ding.
\newblock Self-reflection and articulated consumer preferences.
\newblock {\em Journal of Product Innovation Management}, 31(1):17--32, 2014.

\bibitem[\protect\citeauthoryear{Hejna and Sadigh}{2022}]{iii2022fewshot}
Joseph Hejna and Dorsa Sadigh.
\newblock Few-shot preference learning for human-in-the-loop {RL}.
\newblock In {\em 6th Annual Conference on Robot Learning}, 2022.

\bibitem[\protect\citeauthoryear{Jaccard}{1912}]{jaccard1912distribution}
Paul Jaccard.
\newblock The distribution of the flora in the alpine zone.
\newblock {\em New Phytologist}, 11(2):37--50, 1912.

\bibitem[\protect\citeauthoryear{Kocielnik \bgroup \em et al.\egroup }{2018}]{kocielnik_reflection_2018}
Rafal Kocielnik, Lillian Xiao, Daniel Avrahami, and Gary Hsieh.
\newblock Reflection {Companion}: {A} {Conversational} {System} for {Engaging} {Users} in {Reflection} on {Physical} {Activity}.
\newblock {\em Proceedings of the ACM on Interactive, Mobile, Wearable and Ubiquitous Technologies}, 2(2):70:1--70:26, July 2018.

\bibitem[\protect\citeauthoryear{Kwon \bgroup \em et al.\egroup }{2022}]{kwon2022reward}
Minae Kwon, Sang~Michael Xie, Kalesha Bullard, and Dorsa Sadigh.
\newblock Reward design with language models.
\newblock In {\em The Eleventh International Conference on Learning Representations}, 2022.

\bibitem[\protect\citeauthoryear{Landis and Koch}{1977}]{landis1977measurement}
J~Richard Landis and Gary~G Koch.
\newblock The measurement of observer agreement for categorical data.
\newblock {\em Biometrics}, pages 159--174, 1977.

\bibitem[\protect\citeauthoryear{Le~Dantec \bgroup \em et al.\egroup }{2009}]{le2009values}
Christopher~A Le~Dantec, Erika~Shehan Poole, and Susan~P Wyche.
\newblock Values as lived experience: evolving value sensitive design in support of value discovery.
\newblock In {\em Proceedings of the SIGCHI Conference on Human Factors in Computing Systems}, pages 1141--1150, 2009.

\bibitem[\protect\citeauthoryear{Lichtenstein and Slovic}{2006}]{lichtenstein2006construction}
Sarah Lichtenstein and Paul Slovic.
\newblock The construction of preference: An overview.
\newblock {\em The construction of preference}, 1:1--40, 2006.

\bibitem[\protect\citeauthoryear{Ma \bgroup \em et al.\egroup }{2024}]{ma2024eureka}
Yecheng~Jason Ma, William Liang, Guanzhi Wang, De-An Huang, Osbert Bastani, Dinesh Jayaraman, Yuke Zhu, Linxi Fan, and Anima Anandkumar.
\newblock Eureka: Human-level reward design via coding large language models.
\newblock In {\em The Twelfth International Conference on Learning Representations}, 2024.

\bibitem[\protect\citeauthoryear{Poddar \bgroup \em et al.\egroup }{2024}]{poddar2024personalizing}
Sriyash Poddar, Yanming Wan, Hamish Ivison, Abhishek Gupta, and Natasha Jaques.
\newblock Personalizing reinforcement learning from human feedback with variational preference learning.
\newblock In {\em The Thirty-eighth Annual Conference on Neural Information Processing Systems}, 2024.

\bibitem[\protect\citeauthoryear{Schwartz}{1992}]{schwartz1992universals}
Shalom~H Schwartz.
\newblock Universals in the content and structure of values: Theoretical advances and empirical tests in 20 countries.
\newblock {\em Advances in Experimental Social Psychology/Academic Press}, 1992.

\bibitem[\protect\citeauthoryear{Siththaranjan \bgroup \em et al.\egroup }{2023}]{siththaranjan2023distributional}
Anand Siththaranjan, Cassidy Laidlaw, and Dylan Hadfield-Menell.
\newblock Understanding hidden context in preference learning: Consequences for rlhf.
\newblock In {\em The Twelfth International Conference on Learning Representations}, 2023.

\bibitem[\protect\citeauthoryear{Ver~Donck \bgroup \em et al.\egroup }{2020}]{ver2020improving}
Niki Ver~Donck, Geert Vander~Stichele, Isabelle Huys, et~al.
\newblock Improving patient preference elicitation by applying concepts from the consumer research field: narrative literature review.
\newblock {\em Interactive Journal of Medical Research}, 9(1):e13684, 2020.

\bibitem[\protect\citeauthoryear{Verma \bgroup \em et al.\egroup }{2024}]{verma2024balancing}
Shresth Verma, Niclas Boehmer, Lingkai Kong, and Milind Tambe.
\newblock Balancing act: Prioritization strategies for {LLM}-designed restless bandit rewards.
\newblock In {\em Workshop on Socially Responsible Language Modelling Research}, 2024.

\bibitem[\protect\citeauthoryear{Wolfbauer \bgroup \em et al.\egroup }{2022}]{wolfbauer_script_2022}
Irmtraud Wolfbauer, Viktoria Pammer-Schindler, Katharina Maitz, and Carolyn~P. Rose.
\newblock A {Script} for {Conversational} {Reflection} {Guidance}: {A} {Field} {Study} on {Developing} {Reflection} {Competence} {With} {Apprentices}.
\newblock {\em IEEE Transactions on Learning Technologies}, 15(5):554--566, October 2022.

\bibitem[\protect\citeauthoryear{Xie \bgroup \em et al.\egroup }{2024}]{xie2024textreward}
Tianbao Xie, Siheng Zhao, Chen~Henry Wu, Yitao Liu, Qian Luo, Victor Zhong, Yanchao Yang, and Tao Yu.
\newblock Text2reward: Reward shaping with language models for reinforcement learning.
\newblock In {\em The Twelfth International Conference on Learning Representations}, 2024.

\end{thebibliography}

\end{document}